\documentclass[letterpaper]{article} % DO NOT CHANGE THIS
\usepackage{aaai2026}  % DO NOT CHANGE THIS
\usepackage{times}  % DO NOT CHANGE THIS
\usepackage{helvet}  % DO NOT CHANGE THIS
\usepackage{courier}  % DO NOT CHANGE THIS
\usepackage[hyphens]{url}  % DO NOT CHANGE THIS
\usepackage{graphicx} % DO NOT CHANGE THIS
\usepackage{natbib}  % DO NOT CHANGE THIS AND DO NOT ADD ANY OPTIONS TO IT
\usepackage{caption}  % DO NOT CHANGE THIS AND DO NOT ADD ANY OPTIONS TO IT
\usepackage{amsmath,amssymb}
\usepackage{booktabs}

\providecommand{\Claim}[1]{\csname claim@#1\endcsname}
\expandafter\newcommand\csname claim@fam_post_scenic_entry_recon_scenic\endcsname{0.46}
\expandafter\newcommand\csname claim@fam_post_scenic_entry_recon_scenic_ci_low\endcsname{0.33}
\expandafter\newcommand\csname claim@fam_post_scenic_entry_recon_scenic_ci_high\endcsname{0.59}
\expandafter\newcommand\csname claim@fam_post_scenic_entry_humphrey_scenic\endcsname{0.46}
\expandafter\newcommand\csname claim@fam_post_scenic_entry_humphrey_scenic_ci_low\endcsname{0.33}
\expandafter\newcommand\csname claim@fam_post_scenic_entry_humphrey_scenic_ci_high\endcsname{0.59}
\expandafter\newcommand\csname claim@fam_post_scenic_entry_hb_scenic\endcsname{0.94}
\expandafter\newcommand\csname claim@fam_post_scenic_entry_hb_scenic_ci_low\endcsname{0.83}
\expandafter\newcommand\csname claim@fam_post_scenic_entry_hb_scenic_ci_high\endcsname{1.00}
\expandafter\newcommand\csname claim@fam_post_scenic_entry_recon_dull\endcsname{0.53}
\expandafter\newcommand\csname claim@fam_post_scenic_entry_recon_dull_ci_low\endcsname{0.42}
\expandafter\newcommand\csname claim@fam_post_scenic_entry_recon_dull_ci_high\endcsname{0.65}
\expandafter\newcommand\csname claim@fam_post_scenic_entry_humphrey_dull\endcsname{0.53}
\expandafter\newcommand\csname claim@fam_post_scenic_entry_humphrey_dull_ci_low\endcsname{0.42}
\expandafter\newcommand\csname claim@fam_post_scenic_entry_humphrey_dull_ci_high\endcsname{0.65}
\expandafter\newcommand\csname claim@fam_post_scenic_entry_hb_dull\endcsname{0.96}
\expandafter\newcommand\csname claim@fam_post_scenic_entry_hb_dull_ci_low\endcsname{0.87}
\expandafter\newcommand\csname claim@fam_post_scenic_entry_hb_dull_ci_high\endcsname{1.00}
\expandafter\newcommand\csname claim@fam_delta_scenic_entry_recon\endcsname{0.07}
\expandafter\newcommand\csname claim@fam_delta_scenic_entry_recon_ci_low\endcsname{-0.02}
\expandafter\newcommand\csname claim@fam_delta_scenic_entry_recon_ci_high\endcsname{0.16}
\expandafter\newcommand\csname claim@fam_delta_scenic_entry_humphrey\endcsname{0.07}
\expandafter\newcommand\csname claim@fam_delta_scenic_entry_humphrey_ci_low\endcsname{-0.02}
\expandafter\newcommand\csname claim@fam_delta_scenic_entry_humphrey_ci_high\endcsname{0.16}
\expandafter\newcommand\csname claim@fam_delta_scenic_entry_hb\endcsname{0.01}
\expandafter\newcommand\csname claim@fam_delta_scenic_entry_hb_ci_low\endcsname{0.00}
\expandafter\newcommand\csname claim@fam_delta_scenic_entry_hb_ci_high\endcsname{0.03}
\expandafter\newcommand\csname claim@fam_median_delta_novelty_hb_scenic\endcsname{-0.43}
\expandafter\newcommand\csname claim@fam_valence_scenic_hb\endcsname{0.74}
\expandafter\newcommand\csname claim@fam_valence_scenic_hb_ci_low\endcsname{0.71}
\expandafter\newcommand\csname claim@fam_valence_scenic_hb_ci_high\endcsname{0.77}
\expandafter\newcommand\csname claim@fam_valence_dull_hb\endcsname{0.58}
\expandafter\newcommand\csname claim@fam_valence_dull_hb_ci_low\endcsname{0.58}
\expandafter\newcommand\csname claim@fam_valence_dull_hb_ci_high\endcsname{0.58}
\expandafter\newcommand\csname claim@fam_arousal_scenic_hb\endcsname{0.12}
\expandafter\newcommand\csname claim@fam_arousal_scenic_hb_ci_low\endcsname{0.10}
\expandafter\newcommand\csname claim@fam_arousal_scenic_hb_ci_high\endcsname{0.13}
\expandafter\newcommand\csname claim@fam_arousal_dull_hb\endcsname{0.14}
\expandafter\newcommand\csname claim@fam_arousal_dull_hb_ci_low\endcsname{0.14}
\expandafter\newcommand\csname claim@fam_arousal_dull_hb_ci_high\endcsname{0.14}
\expandafter\newcommand\csname claim@fam_probe_valence_scenic_hb\endcsname{0.91}
\expandafter\newcommand\csname claim@fam_probe_valence_scenic_hb_ci_low\endcsname{0.90}
\expandafter\newcommand\csname claim@fam_probe_valence_scenic_hb_ci_high\endcsname{0.91}
\expandafter\newcommand\csname claim@fam_probe_valence_dull_hb\endcsname{0.88}
\expandafter\newcommand\csname claim@fam_probe_valence_dull_hb_ci_low\endcsname{0.88}
\expandafter\newcommand\csname claim@fam_probe_valence_dull_hb_ci_high\endcsname{0.88}
\expandafter\newcommand\csname claim@fam_probe_arousal_scenic_hb\endcsname{0.24}
\expandafter\newcommand\csname claim@fam_probe_arousal_scenic_hb_ci_low\endcsname{0.22}
\expandafter\newcommand\csname claim@fam_probe_arousal_scenic_hb_ci_high\endcsname{0.26}
\expandafter\newcommand\csname claim@fam_probe_arousal_dull_hb\endcsname{0.32}
\expandafter\newcommand\csname claim@fam_probe_arousal_dull_hb_ci_low\endcsname{0.32}
\expandafter\newcommand\csname claim@fam_probe_arousal_dull_hb_ci_high\endcsname{0.32}
\expandafter\newcommand\csname claim@fam_n_seeds\endcsname{20}
\expandafter\newcommand\csname claim@fam_post_repeats\endcsname{5}
\expandafter\newcommand\csname claim@play_unique_viewpoints_recon\endcsname{136}
\expandafter\newcommand\csname claim@play_unique_viewpoints_recon_ci_low\endcsname{128}
\expandafter\newcommand\csname claim@play_unique_viewpoints_recon_ci_high\endcsname{144}
\expandafter\newcommand\csname claim@play_unique_viewpoints_humphrey\endcsname{140}
\expandafter\newcommand\csname claim@play_unique_viewpoints_humphrey_ci_low\endcsname{133}
\expandafter\newcommand\csname claim@play_unique_viewpoints_humphrey_ci_high\endcsname{147}
\expandafter\newcommand\csname claim@play_unique_viewpoints_hb\endcsname{138}
\expandafter\newcommand\csname claim@play_unique_viewpoints_hb_ci_low\endcsname{124}
\expandafter\newcommand\csname claim@play_unique_viewpoints_hb_ci_high\endcsname{150}
\expandafter\newcommand\csname claim@play_unique_viewpoints_range_low\endcsname{136}
\expandafter\newcommand\csname claim@play_unique_viewpoints_range_high\endcsname{140}
\expandafter\newcommand\csname claim@play_neutral_entropy_recon\endcsname{4.81}
\expandafter\newcommand\csname claim@play_neutral_entropy_recon_ci_low\endcsname{4.77}
\expandafter\newcommand\csname claim@play_neutral_entropy_recon_ci_high\endcsname{4.84}
\expandafter\newcommand\csname claim@play_neutral_entropy_humphrey\endcsname{4.81}
\expandafter\newcommand\csname claim@play_neutral_entropy_humphrey_ci_low\endcsname{4.76}
\expandafter\newcommand\csname claim@play_neutral_entropy_humphrey_ci_high\endcsname{4.85}
\expandafter\newcommand\csname claim@play_neutral_entropy_hb\endcsname{4.57}
\expandafter\newcommand\csname claim@play_neutral_entropy_hb_ci_low\endcsname{4.41}
\expandafter\newcommand\csname claim@play_neutral_entropy_hb_ci_high\endcsname{4.69}
\expandafter\newcommand\csname claim@play_neutral_entropy_range_low\endcsname{4.57}
\expandafter\newcommand\csname claim@play_neutral_entropy_range_high\endcsname{4.81}
\expandafter\newcommand\csname claim@play_internal_actuator_recon_zero\endcsname{True}
\expandafter\newcommand\csname claim@play_internal_actuator_ipsundrum_nontrivial\endcsname{True}
\expandafter\newcommand\csname claim@play_scan_events_recon\endcsname{0.9}
\expandafter\newcommand\csname claim@play_scan_events_recon_ci_low\endcsname{0.6}
\expandafter\newcommand\csname claim@play_scan_events_recon_ci_high\endcsname{1.4}
\expandafter\newcommand\csname claim@play_scan_events_humphrey\endcsname{1.1}
\expandafter\newcommand\csname claim@play_scan_events_humphrey_ci_low\endcsname{0.8}
\expandafter\newcommand\csname claim@play_scan_events_humphrey_ci_high\endcsname{1.4}
\expandafter\newcommand\csname claim@play_scan_events_hb\endcsname{31.4}
\expandafter\newcommand\csname claim@play_scan_events_hb_ci_low\endcsname{25.8}
\expandafter\newcommand\csname claim@play_scan_events_hb_ci_high\endcsname{38.2}
\expandafter\newcommand\csname claim@play_cycle_score_hb\endcsname{7.6}
\expandafter\newcommand\csname claim@play_cycle_score_hb_ci_low\endcsname{1.2}
\expandafter\newcommand\csname claim@play_cycle_score_hb_ci_high\endcsname{16.7}
\expandafter\newcommand\csname claim@play_action_entropy_hb_curiosity\endcsname{1.29}
\expandafter\newcommand\csname claim@play_action_entropy_hb_curiosity_ci_low\endcsname{1.25}
\expandafter\newcommand\csname claim@play_action_entropy_hb_curiosity_ci_high\endcsname{1.33}
\expandafter\newcommand\csname claim@play_action_entropy_random\endcsname{1.99}
\expandafter\newcommand\csname claim@play_action_entropy_random_ci_low\endcsname{1.98}
\expandafter\newcommand\csname claim@play_action_entropy_random_ci_high\endcsname{1.99}
\expandafter\newcommand\csname claim@play_dwell_p90_hb_curiosity\endcsname{2.7}
\expandafter\newcommand\csname claim@play_dwell_p90_hb_curiosity_ci_low\endcsname{2.5}
\expandafter\newcommand\csname claim@play_dwell_p90_hb_curiosity_ci_high\endcsname{2.9}
\expandafter\newcommand\csname claim@play_dwell_p90_random\endcsname{8.97}
\expandafter\newcommand\csname claim@play_dwell_p90_random_ci_low\endcsname{8.29}
\expandafter\newcommand\csname claim@play_dwell_p90_random_ci_high\endcsname{9.77}
\expandafter\newcommand\csname claim@pain_ns_half_life_recon\endcsname{0}
\expandafter\newcommand\csname claim@pain_ns_half_life_recon_ci_low\endcsname{0}
\expandafter\newcommand\csname claim@pain_ns_half_life_recon_ci_high\endcsname{0}
\expandafter\newcommand\csname claim@pain_ns_half_life_humphrey\endcsname{0}
\expandafter\newcommand\csname claim@pain_ns_half_life_humphrey_ci_low\endcsname{0}
\expandafter\newcommand\csname claim@pain_ns_half_life_humphrey_ci_high\endcsname{0}
\expandafter\newcommand\csname claim@pain_ns_half_life_hb\endcsname{0}
\expandafter\newcommand\csname claim@pain_ns_half_life_hb_ci_low\endcsname{0}
\expandafter\newcommand\csname claim@pain_ns_half_life_hb_ci_high\endcsname{0}
\expandafter\newcommand\csname claim@pain_tail_duration_recon\endcsname{5}
\expandafter\newcommand\csname claim@pain_tail_duration_recon_ci_low\endcsname{5}
\expandafter\newcommand\csname claim@pain_tail_duration_recon_ci_high\endcsname{5}
\expandafter\newcommand\csname claim@pain_tail_duration_humphrey\endcsname{5}
\expandafter\newcommand\csname claim@pain_tail_duration_humphrey_ci_low\endcsname{5}
\expandafter\newcommand\csname claim@pain_tail_duration_humphrey_ci_high\endcsname{5}
\expandafter\newcommand\csname claim@pain_tail_duration_hb\endcsname{90}
\expandafter\newcommand\csname claim@pain_tail_duration_hb_ci_low\endcsname{52}
\expandafter\newcommand\csname claim@pain_tail_duration_hb_ci_high\endcsname{128}
\expandafter\newcommand\csname claim@pain_ns_half_life_saturation_recon\endcsname{0.00}
\expandafter\newcommand\csname claim@pain_ns_half_life_saturation_humphrey\endcsname{0.00}
\expandafter\newcommand\csname claim@pain_ns_half_life_saturation_hb\endcsname{0.00}
\expandafter\newcommand\csname claim@lesion_auc_drop_recon\endcsname{0.00}
\expandafter\newcommand\csname claim@lesion_auc_drop_recon_ci_low\endcsname{0.00}
\expandafter\newcommand\csname claim@lesion_auc_drop_recon_ci_high\endcsname{0.00}
\expandafter\newcommand\csname claim@lesion_auc_drop_humphrey\endcsname{19.12}
\expandafter\newcommand\csname claim@lesion_auc_drop_humphrey_ci_low\endcsname{19.12}
\expandafter\newcommand\csname claim@lesion_auc_drop_humphrey_ci_high\endcsname{19.12}
\expandafter\newcommand\csname claim@lesion_auc_drop_hb\endcsname{27.62}
\expandafter\newcommand\csname claim@lesion_auc_drop_hb_ci_low\endcsname{27.60}
\expandafter\newcommand\csname claim@lesion_auc_drop_hb_ci_high\endcsname{27.64}
\expandafter\newcommand\csname claim@lesion_auc_drop_pct_humphrey\endcsname{20.3}
\expandafter\newcommand\csname claim@lesion_auc_drop_pct_humphrey_ci_low\endcsname{20.3}
\expandafter\newcommand\csname claim@lesion_auc_drop_pct_humphrey_ci_high\endcsname{20.3}
\expandafter\newcommand\csname claim@lesion_auc_drop_pct_hb\endcsname{27.9}
\expandafter\newcommand\csname claim@lesion_auc_drop_pct_hb_ci_low\endcsname{27.8}
\expandafter\newcommand\csname claim@lesion_auc_drop_pct_hb_ci_high\endcsname{27.9}
\expandafter\newcommand\csname claim@lesion_recon_auc_drop_near_zero\endcsname{True}
\expandafter\newcommand\csname claim@goal_corridor_hazards_max_hb\endcsname{0.00}
\expandafter\newcommand\csname claim@goal_corridor_hazards_zero_hb\endcsname{True}
\expandafter\newcommand\csname claim@goal_directed_seeds\endcsname{20}
\expandafter\newcommand\csname claim@goal_directed_horizons\endcsname{1,2,3,4,5,6,7,8,9,10,11,12,13,14,15,16,17,18,19,20}
\expandafter\newcommand\csname claim@goal_directed_horizon_count\endcsname{20}
\expandafter\newcommand\csname claim@goal_directed_paper_horizon\endcsname{5}
\expandafter\newcommand\csname claim@goal_corridor_h5_hazards_recon\endcsname{3.05}
\expandafter\newcommand\csname claim@goal_corridor_h5_hazards_recon_ci_low\endcsname{2.00}
\expandafter\newcommand\csname claim@goal_corridor_h5_hazards_recon_ci_high\endcsname{4.25}
\expandafter\newcommand\csname claim@goal_corridor_h5_hazards_humphrey\endcsname{2.90}
\expandafter\newcommand\csname claim@goal_corridor_h5_hazards_humphrey_ci_low\endcsname{1.90}
\expandafter\newcommand\csname claim@goal_corridor_h5_hazards_humphrey_ci_high\endcsname{4.05}
\expandafter\newcommand\csname claim@goal_corridor_h5_hazards_hb\endcsname{0.00}
\expandafter\newcommand\csname claim@goal_corridor_h5_hazards_hb_ci_low\endcsname{0.00}
\expandafter\newcommand\csname claim@goal_corridor_h5_hazards_hb_ci_high\endcsname{0.00}
\expandafter\newcommand\csname claim@goal_corridor_h5_time_recon\endcsname{50.00}
\expandafter\newcommand\csname claim@goal_corridor_h5_time_recon_ci_low\endcsname{37.35}
\expandafter\newcommand\csname claim@goal_corridor_h5_time_recon_ci_high\endcsname{63.20}
\expandafter\newcommand\csname claim@goal_corridor_h5_time_humphrey\endcsname{50.00}
\expandafter\newcommand\csname claim@goal_corridor_h5_time_humphrey_ci_low\endcsname{37.35}
\expandafter\newcommand\csname claim@goal_corridor_h5_time_humphrey_ci_high\endcsname{63.20}
\expandafter\newcommand\csname claim@goal_corridor_h5_time_hb\endcsname{63.15}
\expandafter\newcommand\csname claim@goal_corridor_h5_time_hb_ci_low\endcsname{53.95}
\expandafter\newcommand\csname claim@goal_corridor_h5_time_hb_ci_high\endcsname{71.45}
\expandafter\newcommand\csname claim@goal_corridor_h5_success_recon\endcsname{0.55}
\expandafter\newcommand\csname claim@goal_corridor_h5_success_recon_ci_low\endcsname{0.30}
\expandafter\newcommand\csname claim@goal_corridor_h5_success_recon_ci_high\endcsname{0.75}
\expandafter\newcommand\csname claim@goal_corridor_h5_success_humphrey\endcsname{0.55}
\expandafter\newcommand\csname claim@goal_corridor_h5_success_humphrey_ci_low\endcsname{0.30}
\expandafter\newcommand\csname claim@goal_corridor_h5_success_humphrey_ci_high\endcsname{0.75}
\expandafter\newcommand\csname claim@goal_corridor_h5_success_hb\endcsname{0.45}
\expandafter\newcommand\csname claim@goal_corridor_h5_success_hb_ci_low\endcsname{0.25}
\expandafter\newcommand\csname claim@goal_corridor_h5_success_hb_ci_high\endcsname{0.65}
\expandafter\newcommand\csname claim@goal_gridworld_h5_hazards_recon\endcsname{14.30}
\expandafter\newcommand\csname claim@goal_gridworld_h5_hazards_recon_ci_low\endcsname{8.90}
\expandafter\newcommand\csname claim@goal_gridworld_h5_hazards_recon_ci_high\endcsname{20.05}
\expandafter\newcommand\csname claim@goal_gridworld_h5_hazards_humphrey\endcsname{3.30}
\expandafter\newcommand\csname claim@goal_gridworld_h5_hazards_humphrey_ci_low\endcsname{1.35}
\expandafter\newcommand\csname claim@goal_gridworld_h5_hazards_humphrey_ci_high\endcsname{5.70}
\expandafter\newcommand\csname claim@goal_gridworld_h5_hazards_hb\endcsname{0.15}
\expandafter\newcommand\csname claim@goal_gridworld_h5_hazards_hb_ci_low\endcsname{0.00}
\expandafter\newcommand\csname claim@goal_gridworld_h5_hazards_hb_ci_high\endcsname{0.30}
\expandafter\newcommand\csname claim@goal_gridworld_h5_time_recon\endcsname{171.80}
\expandafter\newcommand\csname claim@goal_gridworld_h5_time_recon_ci_low\endcsname{131.30}
\expandafter\newcommand\csname claim@goal_gridworld_h5_time_recon_ci_high\endcsname{211.10}
\expandafter\newcommand\csname claim@goal_gridworld_h5_time_humphrey\endcsname{135.90}
\expandafter\newcommand\csname claim@goal_gridworld_h5_time_humphrey_ci_low\endcsname{94.95}
\expandafter\newcommand\csname claim@goal_gridworld_h5_time_humphrey_ci_high\endcsname{176.10}
\expandafter\newcommand\csname claim@goal_gridworld_h5_time_hb\endcsname{17.75}
\expandafter\newcommand\csname claim@goal_gridworld_h5_time_hb_ci_low\endcsname{4.35}
\expandafter\newcommand\csname claim@goal_gridworld_h5_time_hb_ci_high\endcsname{43.50}
\expandafter\newcommand\csname claim@goal_gridworld_h5_success_recon\endcsname{0.50}
\expandafter\newcommand\csname claim@goal_gridworld_h5_success_recon_ci_low\endcsname{0.30}
\expandafter\newcommand\csname claim@goal_gridworld_h5_success_recon_ci_high\endcsname{0.70}
\expandafter\newcommand\csname claim@goal_gridworld_h5_success_humphrey\endcsname{0.70}
\expandafter\newcommand\csname claim@goal_gridworld_h5_success_humphrey_ci_low\endcsname{0.50}
\expandafter\newcommand\csname claim@goal_gridworld_h5_success_humphrey_ci_high\endcsname{0.90}
\expandafter\newcommand\csname claim@goal_gridworld_h5_success_hb\endcsname{0.95}
\expandafter\newcommand\csname claim@goal_gridworld_h5_success_hb_ci_low\endcsname{0.85}
\expandafter\newcommand\csname claim@goal_gridworld_h5_success_hb_ci_high\endcsname{1.00}
\expandafter\newcommand\csname claim@goal_corridor_all_hazards_recon\endcsname{3.05}
\expandafter\newcommand\csname claim@goal_corridor_all_hazards_recon_ci_low\endcsname{2.00}
\expandafter\newcommand\csname claim@goal_corridor_all_hazards_recon_ci_high\endcsname{4.25}
\expandafter\newcommand\csname claim@goal_corridor_all_hazards_humphrey\endcsname{2.90}
\expandafter\newcommand\csname claim@goal_corridor_all_hazards_humphrey_ci_low\endcsname{1.90}
\expandafter\newcommand\csname claim@goal_corridor_all_hazards_humphrey_ci_high\endcsname{4.05}
\expandafter\newcommand\csname claim@goal_corridor_all_hazards_hb\endcsname{0.00}
\expandafter\newcommand\csname claim@goal_corridor_all_hazards_hb_ci_low\endcsname{0.00}
\expandafter\newcommand\csname claim@goal_corridor_all_hazards_hb_ci_high\endcsname{0.00}
\expandafter\newcommand\csname claim@goal_corridor_all_time_recon\endcsname{50.00}
\expandafter\newcommand\csname claim@goal_corridor_all_time_recon_ci_low\endcsname{37.35}
\expandafter\newcommand\csname claim@goal_corridor_all_time_recon_ci_high\endcsname{63.20}
\expandafter\newcommand\csname claim@goal_corridor_all_time_humphrey\endcsname{49.73}
\expandafter\newcommand\csname claim@goal_corridor_all_time_humphrey_ci_low\endcsname{37.12}
\expandafter\newcommand\csname claim@goal_corridor_all_time_humphrey_ci_high\endcsname{62.83}
\expandafter\newcommand\csname claim@goal_corridor_all_time_hb\endcsname{52.71}
\expandafter\newcommand\csname claim@goal_corridor_all_time_hb_ci_low\endcsname{47.16}
\expandafter\newcommand\csname claim@goal_corridor_all_time_hb_ci_high\endcsname{58.40}
\expandafter\newcommand\csname claim@goal_corridor_all_success_recon\endcsname{0.55}
\expandafter\newcommand\csname claim@goal_corridor_all_success_recon_ci_low\endcsname{0.30}
\expandafter\newcommand\csname claim@goal_corridor_all_success_recon_ci_high\endcsname{0.75}
\expandafter\newcommand\csname claim@goal_corridor_all_success_humphrey\endcsname{0.58}
\expandafter\newcommand\csname claim@goal_corridor_all_success_humphrey_ci_low\endcsname{0.35}
\expandafter\newcommand\csname claim@goal_corridor_all_success_humphrey_ci_high\endcsname{0.78}
\expandafter\newcommand\csname claim@goal_corridor_all_success_hb\endcsname{0.73}
\expandafter\newcommand\csname claim@goal_corridor_all_success_hb_ci_low\endcsname{0.61}
\expandafter\newcommand\csname claim@goal_corridor_all_success_hb_ci_high\endcsname{0.83}
\expandafter\newcommand\csname claim@goal_gridworld_all_hazards_recon\endcsname{14.30}
\expandafter\newcommand\csname claim@goal_gridworld_all_hazards_recon_ci_low\endcsname{8.90}
\expandafter\newcommand\csname claim@goal_gridworld_all_hazards_recon_ci_high\endcsname{20.05}
\expandafter\newcommand\csname claim@goal_gridworld_all_hazards_humphrey\endcsname{2.45}
\expandafter\newcommand\csname claim@goal_gridworld_all_hazards_humphrey_ci_low\endcsname{1.27}
\expandafter\newcommand\csname claim@goal_gridworld_all_hazards_humphrey_ci_high\endcsname{3.85}
\expandafter\newcommand\csname claim@goal_gridworld_all_hazards_hb\endcsname{0.26}
\expandafter\newcommand\csname claim@goal_gridworld_all_hazards_hb_ci_low\endcsname{0.10}
\expandafter\newcommand\csname claim@goal_gridworld_all_hazards_hb_ci_high\endcsname{0.44}
\expandafter\newcommand\csname claim@goal_gridworld_all_time_recon\endcsname{171.80}
\expandafter\newcommand\csname claim@goal_gridworld_all_time_recon_ci_low\endcsname{131.30}
\expandafter\newcommand\csname claim@goal_gridworld_all_time_recon_ci_high\endcsname{211.10}
\expandafter\newcommand\csname claim@goal_gridworld_all_time_humphrey\endcsname{129.77}
\expandafter\newcommand\csname claim@goal_gridworld_all_time_humphrey_ci_low\endcsname{90.63}
\expandafter\newcommand\csname claim@goal_gridworld_all_time_humphrey_ci_high\endcsname{168.70}
\expandafter\newcommand\csname claim@goal_gridworld_all_time_hb\endcsname{9.54}
\expandafter\newcommand\csname claim@goal_gridworld_all_time_hb_ci_low\endcsname{4.75}
\expandafter\newcommand\csname claim@goal_gridworld_all_time_hb_ci_high\endcsname{16.23}
\expandafter\newcommand\csname claim@goal_gridworld_all_success_recon\endcsname{0.50}
\expandafter\newcommand\csname claim@goal_gridworld_all_success_recon_ci_low\endcsname{0.30}
\expandafter\newcommand\csname claim@goal_gridworld_all_success_recon_ci_high\endcsname{0.70}
\expandafter\newcommand\csname claim@goal_gridworld_all_success_humphrey\endcsname{0.72}
\expandafter\newcommand\csname claim@goal_gridworld_all_success_humphrey_ci_low\endcsname{0.53}
\expandafter\newcommand\csname claim@goal_gridworld_all_success_humphrey_ci_high\endcsname{0.90}
\expandafter\newcommand\csname claim@goal_gridworld_all_success_hb\endcsname{0.99}
\expandafter\newcommand\csname claim@goal_gridworld_all_success_hb_ci_low\endcsname{0.97}
\expandafter\newcommand\csname claim@goal_gridworld_all_success_hb_ci_high\endcsname{1.00}
\expandafter\newcommand\csname claim@headline_seeds\endcsname{20}
\expandafter\newcommand\csname claim@pain_post_steps\endcsname{200}

\title{ReCoN-Ipsundrum: An Inspectable Recurrent Persistence Loop Agent with Affect-Coupled Control and Mechanism-Linked Consciousness Indicator Assays}
\author{Aishik Sanyal}
\affiliations{Independent Research Engineer\\
aishik@xcellect.com}

\begin{document}
\maketitle

\begin{abstract}
Indicator-based approaches to machine consciousness recommend mechanism-linked evidence triangulated across tasks, supported by architectural inspection and causal intervention.
Inspired by Humphrey's ipsundrum hypothesis, we implement \textbf{ReCoN-Ipsundrum}, an inspectable agent that extends a ReCoN state machine with a recurrent persistence loop over sensory salience $N^s$ and an optional affect proxy reporting valence/arousal.
Across fixed-parameter ablations (ReCoN, Ipsundrum, Ipsundrum+affect), we operationalize Humphrey's \emph{qualiaphilia} (preference for sensory experience for its own sake) as a familiarity-controlled scenic-over-dull route choice. We find a novelty dissociation: non-affect variants are novelty-sensitive ($\Delta\text{scenic-entry}=\Claim{fam_delta_scenic_entry_recon}$). Affect coupling is stable ($\Delta\text{scenic-entry}=\Claim{fam_delta_scenic_entry_hb}$) even when scenic is less novel (median $\Delta\text{novelty}\approx \Claim{fam_median_delta_novelty_hb_scenic}$).
In reward-free \emph{exploratory play}, the affect variant shows structured local investigation (scan events \Claim{play_scan_events_hb} vs.\ \Claim{play_scan_events_recon}; cycle score \Claim{play_cycle_score_hb}).
In a pain-tail probe, only the affect variant sustains prolonged planned caution (tail duration \Claim{pain_tail_duration_hb} vs.\ \Claim{pain_tail_duration_recon}).
Lesioning feedback+integration selectively reduces post-stimulus persistence in ipsundrum variants (AUC drop \Claim{lesion_auc_drop_hb}, \Claim{lesion_auc_drop_pct_hb}\%) while leaving ReCoN unchanged.
These dissociations link recurrence$\rightarrow$persistence and affect-coupled control$\rightarrow$preference stability, scanning, and lingering caution, illustrating how indicator-like signatures can be engineered and why mechanistic and causal evidence should accompany behavioral markers.
\end{abstract}
\begin{links}
    \link{Colab}{https://bit.ly/recips-notebook}
    \link{Code}{https://github.com/xcellect/recips}
    % \link{Datasets}{https://aaai.org/example/datasets}
    % \link{Extended version}{https://aaai.org/example/extended-version}
\end{links}

\section{Introduction}
Recent advances in AI, especially large language models (LLMs) and the post-training procedures that shape their behavior, have revived questions about \emph{machine consciousness} that go beyond intelligence and functional competence.
But ``consciousness'' is a cluster of targets (phenomenal experience, self-in-a-world modeling, robust understanding), and impressive behavior alone does not reveal the internal processes that would make experience \emph{matter}.

Indicator-based approaches treat correlates of consciousness as \emph{credence-shifting} evidence, not decisive tests, and urge triangulation across indicators and evidence types.
This is especially important in AI, where behavior can be achieved by alien mechanisms and by ``minimal'' (gameable) implementations \citep{ButlinEtAl2025Indicators}.
Accordingly, we emphasize \emph{mechanism-linked} tests tied to implementable hypotheses about internal organization, and we report internal signals alongside behavior.

Here we take a constrained step toward mechanism-linked assays for interaction-shaped internal loops.
We implement a small mechanism inspired by Humphrey's theory of sentience: an \emph{ipsundrum}, a closed sensorimotor--interoceptive loop whose recurrent dynamics sustain and ``thicken'' sensation over time \citep{Humphrey2023Sentience}.
We embed ipsundrum dynamics inside a Request Confirmation Network (ReCoN) state machine \citep{BachHerger2015ReCoN} and optionally couple the loop to constructionist affect variables (valence/arousal) \citep{Barrett2017HowEmotions}.
We then test Humphrey-inspired probes by operationalizing \emph{qualiaphilia} as a familiarity-controlled novelty competition to separate hedonic preference from novelty seeking and \emph{exploratory play} via scan events.
Additionally, we evaluate goal-directed navigation as a competence/safety check, a pain-tail assay, and in-episode causal lesions probing post-stimulus sensory persistence.

\paragraph{Terminology and claims.}
Following Butlin et~al., an \emph{indicator} is a theory-derived feature that can raise or lower credence (not a decisive criterion); a \emph{marker} is its measurable operationalization in our assays.
We do \emph{not} claim that any agent here is conscious; we provide an inspectable architecture and falsifiable tests.

\paragraph{Contributions.}
\begin{itemize}
\item \textbf{ReCoN-Ipsundrum:} A ReCoN extension with explicit ipsundrum recurrence and optional affect/interoception coupling \citep{BachHerger2015ReCoN,Humphrey2023Sentience,Barrett2017HowEmotions}.
\item \textbf{Assays with controls:} Operationalized qualiaphilia with a familiarity-controlled regime that adds novelty competition via cross-episode visitation memory, exploratory play with scan events and a pain-tail probe.
\item \textbf{Mechanistic evaluation:} Behavioral and internal measures (e.g., post-stimulus sensory persistence) and within-episode causal lesions that selectively remove recurrence to test mechanism-linked predictions.
\end{itemize}

\section{Related Work}
In the spirit of \emph{haptic realism}, much of what we call understanding is forged through interaction \citep{Chirimuuta2024BrainAbstracted,Chang2022Realism}.
Animals learn by acting, sensing consequences, and regulating internal needs; a purely linguistic model risks a Plato's-cave relation to the world.
Many theories (and the indicator landscape) treat agency, recurrence, predictive regulation, and forms of embodiment as key background conditions \citep{ButlinEtAl2025Indicators}.

A useful parallel comes from skill learning in humans: mental rehearsal can help, but it is not a replacement for real practice and is typically best when paired with it \citep{SchusterEtAl2011BestPracticeMI,LaddaEtAl2021MotorImageryReview}.
Modern AI training similarly mixes ``offline'' optimization (pretraining and fine-tuning) with outcome-based post-training (e.g., RLHF or verifiable-reward RL), often via parameter-efficient adaptation such as LoRA \citep{OuyangEtAl2022InstructGPT,ShaoEtAl2024DeepSeekMath,HuEtAl2021LoRA}.
These methods can yield large \emph{task} gains yet still primarily drive specialization rather than general intelligence \citep{Schulman2025LoRAWithoutRegret}, and they remain ``mental'' unless the system is coupled to ongoing sensorimotor and interoceptive interaction.
\paragraph{Reflex theory as a simplifying ideal.}
Sensorimotor loop models have long served as productive ideals in neuroscience: conditioning framed behavior in terms of modifiable reflexes \citep{Pavlov1927ConditionedReflexes}. Sherrington's ``reflex theory'' \citep{Sherrington1906Integrative} 
treated reflex-arc as a unit mechanism for nervous function. At the same time, Sherrington and later commentators emphasized that the ``simple reflex'' is an analytic abstraction which can mislead if treated as complete \citep{Chirimuuta2024BrainAbstracted}. We therefore start from an explicitly sensorimotor backbone and then test what changes when additional, theory-motivated neurophysiological functions are added and lesioned.

\paragraph{ReCoN and active perception.}
ReCoN is a spreading-activation, message-passing neurosymbolic architecture for executing sensorimotor scripts: routines request confirmation from subordinate routines/sensors and propagate confirmation, inhibition, failure, activations etc. states through a structured graph \citep{BachHerger2015ReCoN}.
It supports an action-to-confirmation view of perception \citep{OReganNoe2001Sensorimotor}, but does not by itself claim to implement the persistent privatized feedback and attractor-like dynamics emphasized by some sentience theories.
However, the neural definition of ReCoN enables learning which, when scaled up, could potentially yield a form of recurrent moment of nowness with second order perception \citep{BachHerger2015ReCoN,Bach2009PSI,Varela1999SpeciousPresent,Lamme2006NeuralStance,LauRosenthal2011EmpiricalHOT,Cleeremans2011RadicalPlasticity}.
% cite: Bach, J. (2009). Principles of Synthetic Intelligence: PSI: An Architecture for Motivated Agents. Oxford University Press.
We will leave this as an open question for future work.

\paragraph{From scripts to sustained loops.}
Our extension strategy keeps ReCoN's clean execution backbone and adds (then lesions) additional causal structure: (i) Humphrey-style monitored sensation and ipsundrum recurrence and (ii) optional affect/interoception coupling.
The goal is not to equate any mechanism with sentience, but to generate testable predictions for \emph{candidate} indicators under explicit~hypotheses.

\paragraph{Ipsundrum, affect, and indicator methodology.}
Humphrey proposes that sentience emerges when reflexive control becomes self-monitoring and then self-sustaining, yielding an ipsundrum attractor and motivating probes like qualiaphilia and exploratory play \citep{Humphrey2023Sentience}.
Constructionist and predictive-processing traditions emphasize affect/interoception and prediction-evaluation loops \citep{Barrett2017HowEmotions,Friston2010FreeEnergy,WolpertEtAl1995InternalModel}, motivating our optional affect coupling.
Finally, Butlin et~al.\ stress that AI assessments should triangulate theory-derived indicators and include architectural/causal evidence because behavior is gameable \citep{ButlinEtAl2025Indicators}; we follow this by mapping mechanisms to indicators (Table~\ref{tab:butlin_coverage}) and using causal lesions that remove recurrence.
\begin{table}[t]
\centering
{\small
\begin{tabular}{@{}p{0.21\linewidth}p{0.48\linewidth}p{0.21\linewidth}@{}}
\toprule
Indicator signpost (Butlin et al.) & What we implement (minimal correspondence) & Important gaps / non-claims \\
\midrule
RPT-1 (recurrent processing) & A localized recurrent loop sustaining $N^s$ after transient stimulus; explicit lesion of feedback+integration. & Not a neural RPT model; no learning; toy domains. \\
PP-1 (predictive processing) & Short-horizon internal rollout using a one-step forward model; epistemic term based on predicted sensory change. & Not hierarchical predictive coding; no online learning; not full active inference. \\
Interoception affect (broadly) & A body-budget proxy and valence/arousal readouts; optional modulation of gain/precision by affect. & Not physiological sensing; no rich autonomic loop; abstraction only. \\
\bottomrule
\end{tabular}
}
\caption{We use selected indicator labels from \citet{ButlinEtAl2025Indicators} only as \emph{design signposts}. ``Correspondence'' denotes a minimal computational element, not a realization of the full theory family.}
\label{tab:butlin_coverage}
\end{table}

\section{Model Summary}
\subsection{ReCoN Substrate: Message-Passing Scripts as Reflexive Sensorimotor Control}
Our agent is built on a small Request Confirmation Network (ReCoN) \citep{BachHerger2015ReCoN}, implemented as an explicit message-passing state machine (\path{core/recon_core.py}, \path{core/recon_network.py}). We do not implement the full MicroPsi architecture of ReCoN with neural learning for inspectability and scope of this paper.
Nodes are typed as \emph{scripts}, \emph{sensors}, or \emph{actuators}. Script nodes occupy a discrete finite state (\path{inactive/requested/active/confirmed/failed}) and exchange \path{request/confirm/wait/inhibit} messages; sensor/actuator nodes expose continuous activations with thresholded confirmation.
This gives a transparent substrate for hierarchical (\path{sub/sur}) ``scripts'' (top-down requests) and confirmation (bottom-up evidence). The models evaluated in this paper do not utilize the sequential (\path{por/ret}) connections.

\subsection{Humphrey Stages as Staged Extensions on the ReCoN Substrate}
Humphrey’s chapter~12 (“The Road Taken”) in \emph{Sentience: The Invention of
  Consciousness} sketches an evolutionary route from approach/avoid reflexes (\emph{sentition})
  to private \emph{sensation} \citep{Humphrey2023Sentience}. As control
  centralizes in a ganglion, an \emph{efference copy} of the motor command
  supports monitoring and “meaning.” When action becomes maladaptive, the
  command is “privatized,” retargeted to an internal body map. In complex
  brains, sensory input couples to this internalized program in a re-entrant loop
  that can stabilize into a repeating attractor (the \emph{ipsundrum}). It
  highlights commandeered motor commands, privatization, and feedback-driven
  attractors.

We treat this narrative as a design scaffold rather than a biological claim.
Operationally, our staged constructors in \path{core/ipsundrum_model.py} begin at the centrally coordinated reflex point (we do not model a purely local, surface-organized reflex):
\textbf{Stage A} implements centrally coordinated reflex sentition as a minimal reflex script $\text{Root}\rightarrow R$ with a sensory terminal $N^s$ and a motor-command proxy $N^m$.
\textbf{Stage B} adds an explicit efference-copy sensor $N^e$, implemented as a low-pass filtered copy of outgoing motor-command magnitude (a monitorable ``what I'm doing'' signal).
\textbf{Stage C} privatizes and thickens sentition by attaching a recurrent ipsundrum state update and forcing the percept script node ($P$) to loop internally for a fixed number of cycles.
\textbf{Stage D} adds a simple gating rule: the percept script continues looping while $N^e$ remains above a threshold, yielding an attractor-like settling regime. The ReCoN network corresponding
to each stage can be viewed in the supplementary Colab notebook.

\subsection{Evaluated Variants (Fixed-Parameter Ablations)}
We evaluate three fixed-parameter variants (no learning). They share the same policy and environment interface; they differ only in internal dynamics and which internal variables exist:
\begin{table}[t]
\centering
{\small
\begin{tabular}{@{}p{0.50\columnwidth}ccc@{}}
\toprule
Model & \shortstack{Ipsundrum\\recurrence} & \shortstack{Affect\\proxy} & \shortstack{Gated\\$P$} \\
\midrule
ReCoN (baseline; stage B) & \texttimes & \texttimes & \texttimes \\
Ipsundrum (stage D) & \checkmark & \texttimes & \checkmark \\
Ipsundrum+affect (stage D) & \checkmark & \checkmark & \checkmark \\
\bottomrule
\end{tabular}
}
\caption{Model variants evaluated. The two ipsundrum rows differ only by the affect proxy layer (and its optional coupling to loop gain/precision).}
\label{tab:model_variants}
\end{table}

\subsection{Sensory Drive $I_t$ and Terminal Semantics ($N^s$, $N^e$)}
Each environment step produces a signed sensory-evidence scalar $I_t\in[-1,1]$ by fusing touch, smell, and vision-cone features (\texttt{core/driver/sensory.py}).
Positive $I_t$ corresponds to noxious evidence (hazard/contact/aversive cues) and negative $I_t$ to scenic/beneficial evidence (beauty cues); importantly, $I_t$ is \emph{not} an external reward.

The primary terminal for Humphrey-style ``sentition'' is $N^s$ (implemented as sensor \texttt{Ns}), which the policy treats as a salience/cost-like internal variable (higher predicted $N^s$ reduces action score; Eq.~\ref{eq:score}).
Crucially for stage separation, when affect is \emph{disabled} we rectify negative input ($I_t\leftarrow \max(0,I_t)$), so non-affect variants can represent \emph{cost} but do not obtain a built-in ``pleasantness'' benefit from negative $I_t$.

Stage B introduces $N^e$ (sensor \texttt{Ne}) as an efference copy: a low-pass filtered copy of the magnitude of the outgoing motor command proxy. This provides a monitorable, temporally extended signal without introducing ipsundrum recurrence.

\subsection{Stage C/D: Ipsundrum Dynamics as a Single-Step State Update}
Stage C and D attach a recurrent ``ipsundrum'' state update to the ReCoN terminals. The implementation is a pure function \texttt{ipsundrum\_step} (\path{core/driver/ipsundrum_dynamics.py}) used \emph{both} online and in the forward model for planning, ensuring ablation fidelity.

Let $E_{t-1}$ be the previous reafferent signal, $\pi_t$ an effective precision, and $b$ a bias term (we set \texttt{sensor\_bias}=0.5 to map signed $I_t$ into the $[0,1]$ sensor range). The update is:
\begin{align}
\text{drive}_t &= I_t + \pi_t\,E_{t-1} + b + \epsilon_t,\\
N^s_t &= \mathrm{clip}_{[0,1]}\!\left(F(\text{drive}_t)\right),
\end{align}
where $F$ is a chosen nonlinearity (linear or sigmoid) and $\epsilon_t$ optional noise (set to zero in our headline runs).
A ``thick-moment'' integrator produces persistence:
\begin{align}
X_t &= d\,X_{t-1} + (1-d)\,N^s_t,\\
M_t &= \mathrm{clip}_{[0,1]}(h\,X_t),\\
N^e_t &= d_e\,N^e_{t-1} + (1-d_e)\,M_t,\\
E_t &= \mathrm{clip}_{[0,1]}(g_{\text{eff}}\,M_t).
\end{align}
We include optional fatigue and divisive normalization terms in code to avoid hard saturation under strong feedback; full parameters are exported in \texttt{results/paper/params\_table.tex}.
The lesion assay uses explicit flags that zero out feedback ($E_{t-1}\!=\!0$ and $\pi_t\!=\!0$) and/or bypass integration ($d\!=\!0$), allowing within-episode causal attribution.

A convenient diagnostic is the \emph{effective recurrence strength} reported by the agent:
\begin{equation}
\alpha_{\text{eff}}\;=\; d + (1-d)\,(g_{\text{eff}}\,h\,\pi_t),
\end{equation}
which distinguishes passive decay from actively maintained recurrence.

\subsection{Stage D+: Barrett-Style Affect as an Interoceptive Proxy}
The affect extension (\texttt{AffectParams} in \texttt{core/ipsundrum\_model.py}) implements a minimal ``body-budget'' proxy and readouts inspired by constructionist affect \citep{Barrett2017HowEmotions}.
An internal budget model $bb_t$ is updated by prediction error under a homeostatic controller, with a \emph{signed} stimulus demand term: positive $I_t$ contributes cost (depleting the budget) while negative $I_t$ contributes deposit (replenishing the budget).
We expose:
(i) an interoceptive proxy sensor $N^i$ (budget model), and
(ii) readouts $N^v$ (valence: closeness to setpoint) and $N^a$ (arousal: magnitude of prediction error and demand).
These nodes exist only when affect is enabled; otherwise they are absent and treated as undefined in logs (NaN).

When enabled, affect can also modulate ipsundrum parameters (precision and/or feedback gain) as a simple form of affect-coupled control, thereby changing $\alpha_{\text{eff}}$ in a state-dependent way.

\subsection{Policy: Short-Horizon Internal Rollout with Model-Aligned Forward Dynamics}
All variants use the same action-selection routine (\path{core/driver/active_perception.py}): enumerate actions, simulate their sensory consequences, and evaluate short-horizon internal rollouts using a one-step forward model.
The forward model is variant-aligned: ReCoN planning uses \texttt{predict\_one\_step\_recon} (no ipsundrum state), while ipsundrum variants use \texttt{ipsundrum\_step} via \path{core/driver/ipsundrum_forward.py}.
Tie-breaking is deterministic given the seed (we shuffle candidate-action order using the episode RNG).

Let $N^v$ and $N^a$ be predicted valence/arousal when affect is enabled, $N^s$ predicted salience, and $bb$ the predicted budget-model state with setpoint $sp$.
The base internal score for an action is:
\begin{equation}
\label{eq:score}
\begin{split}
\mathrm{Score} \;=&\;\underbrace{w_v N^v + w_a N^a + w_s N^s + w_{bb}|bb-sp|}_{\text{affect/regulation}}\\
&+\underbrace{w_{\mathrm{epi}}\,|I_{\mathrm{pred}}-I_{\mathrm{cur}}|}_{\text{epistemic}}\\
&+\underbrace{\mathrm{novelty\ bonus}}_{\text{curiosity}}\\
&+\underbrace{w_{\mathrm{prog}}\cdot\mathrm{progress}}_{\text{goal progress}}\\
&-\underbrace{w_{\mathrm{haz}}\cdot I_{\mathrm{touch,pred}}}_{\text{hazard penalty}}\\
&-\underbrace{\mathrm{action\ costs}}_{\text{action cost}}.
\end{split}
\end{equation}
In our headline affect-coupled runs, $(w_v,w_a,w_s,w_{bb})=(2.0,-1.2,-0.8,-0.4)$; the ReCoN baseline sets these to zero (pure ``script+planning'' substrate).
Additional small terms in code implement a low-change epistemic penalty, a mild forward prior, an arousal-gated caution penalty for forward moves (to link high arousal to avoidance), and a small hazard-touch penalty proportional to the predicted touch sensor ($I_{\mathrm{touch,pred}}$; default scale $w_{\mathrm{haz}}{=}0.10$).
We emphasize transparency because it determines construct validity: in corridor assays we disable an explicit beauty term, but scenic vs.\ dull still changes $I_t$ and therefore the internal variables that enter Eq.~\ref{eq:score}.

\section{Assays, Analysis, and Results}
Unless stated otherwise, uncertainty intervals are 95\% bootstrap confidence intervals of the \emph{per-seed} mean (2000 resamples; seeds are the independent unit).
All quantitative results below are computed from the released \texttt{results/} artifacts in the provided codebase.

\begin{table}[t]
\centering
{\small
\begin{tabular}{@{}p{0.21\linewidth}p{0.46\linewidth}p{0.22\linewidth}@{}}
\toprule
Assay & Setup / manipulation & Primary readouts \\
\midrule
Goal-directed navigation & CorridorWorld and GridWorld with hazards; explicit progress term enabled; curiosity off; sweep planning horizon $H$. & Hazard contacts; steps-to-goal; success rate. \\
Corridor preference with familiarity control & Two equally safe routes (scenic vs.\ dull). Pre-exposure manipulates lane familiarity; post episodes add a curiosity bonus from visitation memory. & Scenic-entry rate; novelty sensitivity $\Delta$scenic-entry; (Ipsundrum+affect) internal valence/arousal probe. \\
Exploratory play & Reward-free neutral-texture gridworld (200 steps), with curiosity enabled in the headline play condition. & Unique viewpoints; scan events; cycle score; action entropy/dwell. \\
Pain-tail & Force one hazard contact, then \emph{remove} the hazard and hold the state fixed while recording \emph{planned} actions for \Claim{pain_post_steps} steps. & Post-stimulus $N^s$ AUC above baseline; planned turn-rate tail duration. \\
Causal lesion & In-episode lesion at $t{=}3$ disabling ipsundrum feedback+integration (vs.\ sham). & Post-lesion $N^s$ AUC and AUC drop (150-step window). \\
\bottomrule
\end{tabular}
}
\caption{Assays used in this paper and their primary readouts.}
\label{tab:assays}
\end{table}

\subsection{Goal-Directed Navigation (Competence and Safety Check)}
We evaluate CorridorWorld and GridWorld navigation under a progress-augmented objective (Eq.~\ref{eq:score}) with curiosity disabled, sweeping rollout horizons $H\in\{1,\dots,20\}$.
Table~\ref{tab:goalperf} aggregates hazard contacts, steps-to-goal, and success.
Ipsundrum+affect reduces hazard contacts in both environments; in GridWorld this also improves success and time-to-goal, while in CorridorWorld it remains hazard-free but slower.
These tasks are competence/safety checks, not consciousness indicators.
Table~\ref{tab:dissociation} previews how our three variants dissociate across the indicator-like signatures we probe; the subsections below unpack each assay.

\begin{table*}[t]
\centering

{\small
\setlength{\tabcolsep}{1.0mm}
\begin{tabular}{lccc|ccc}
\hline
& \multicolumn{3}{c|}{CorridorWorld (all horizons)} & \multicolumn{3}{c}{GridWorld (all horizons)} \\
Model & Hazards & Time & Success & Hazards & Time & Success \\
\hline
ReCoN &
\Claim{goal_corridor_all_hazards_recon}[\Claim{goal_corridor_all_hazards_recon_ci_low},\Claim{goal_corridor_all_hazards_recon_ci_high}] &
\Claim{goal_corridor_all_time_recon}[\Claim{goal_corridor_all_time_recon_ci_low},\Claim{goal_corridor_all_time_recon_ci_high}] &
\Claim{goal_corridor_all_success_recon}[\Claim{goal_corridor_all_success_recon_ci_low},\Claim{goal_corridor_all_success_recon_ci_high}] &
\Claim{goal_gridworld_all_hazards_recon}[\Claim{goal_gridworld_all_hazards_recon_ci_low},\Claim{goal_gridworld_all_hazards_recon_ci_high}] &
\Claim{goal_gridworld_all_time_recon}[\Claim{goal_gridworld_all_time_recon_ci_low},\Claim{goal_gridworld_all_time_recon_ci_high}] &
\Claim{goal_gridworld_all_success_recon}[\Claim{goal_gridworld_all_success_recon_ci_low},\Claim{goal_gridworld_all_success_recon_ci_high}] \\
Ipsundrum &
\Claim{goal_corridor_all_hazards_humphrey}[\Claim{goal_corridor_all_hazards_humphrey_ci_low},\Claim{goal_corridor_all_hazards_humphrey_ci_high}] &
\Claim{goal_corridor_all_time_humphrey}[\Claim{goal_corridor_all_time_humphrey_ci_low},\Claim{goal_corridor_all_time_humphrey_ci_high}] &
\Claim{goal_corridor_all_success_humphrey}[\Claim{goal_corridor_all_success_humphrey_ci_low},\Claim{goal_corridor_all_success_humphrey_ci_high}] &
\Claim{goal_gridworld_all_hazards_humphrey}[\Claim{goal_gridworld_all_hazards_humphrey_ci_low},\Claim{goal_gridworld_all_hazards_humphrey_ci_high}] &
\Claim{goal_gridworld_all_time_humphrey}[\Claim{goal_gridworld_all_time_humphrey_ci_low},\Claim{goal_gridworld_all_time_humphrey_ci_high}] &
\Claim{goal_gridworld_all_success_humphrey}[\Claim{goal_gridworld_all_success_humphrey_ci_low},\Claim{goal_gridworld_all_success_humphrey_ci_high}] \\
Ipsundrum+affect &
\Claim{goal_corridor_all_hazards_hb}[\Claim{goal_corridor_all_hazards_hb_ci_low},\Claim{goal_corridor_all_hazards_hb_ci_high}] &
\Claim{goal_corridor_all_time_hb}[\Claim{goal_corridor_all_time_hb_ci_low},\Claim{goal_corridor_all_time_hb_ci_high}] &
\Claim{goal_corridor_all_success_hb}[\Claim{goal_corridor_all_success_hb_ci_low},\Claim{goal_corridor_all_success_hb_ci_high}] &
\Claim{goal_gridworld_all_hazards_hb}[\Claim{goal_gridworld_all_hazards_hb_ci_low},\Claim{goal_gridworld_all_hazards_hb_ci_high}] &
\Claim{goal_gridworld_all_time_hb}[\Claim{goal_gridworld_all_time_hb_ci_low},\Claim{goal_gridworld_all_time_hb_ci_high}] &
\Claim{goal_gridworld_all_success_hb}[\Claim{goal_gridworld_all_success_hb_ci_low},\Claim{goal_gridworld_all_success_hb_ci_high}] \\
\hline
\end{tabular}
}
\caption{Goal-directed performance aggregated across the horizon sweep (per-seed means over horizons; 95\% bootstrap CI over $N=\Claim{goal_directed_seeds}$ seeds). Time is steps-to-goal (failures counted at the time limit).}
\label{tab:goalperf}
\end{table*}

% --- Signature/component dissociation overview (kept adjacent to Table~\ref{tab:goalperf}) ---
\begin{table*}[t]
\centering
{\small
\setlength{\tabcolsep}{4pt}
\renewcommand{\arraystretch}{1.05}
\begin{tabular}{@{}p{0.62\textwidth}ccc@{}}
\toprule
Signature (assay) & Recon & Ipsundrum & Ipsundrum+affect \\
\midrule
Persistence in $N^s$ (lesion / pain-tail AUC) & \ensuremath{\times} & \ensuremath{\checkmark} & \ensuremath{\checkmark} \\
Valence-stable scenic preference (familiarity control) & \ensuremath{\times} & \ensuremath{\times} & \ensuremath{\checkmark} \\
Structured local scanning (play: scan events) & \ensuremath{\times} & \ensuremath{\times} & \ensuremath{\checkmark} \\
Lingering planned caution (pain-tail: tail duration) & \ensuremath{\times} & \ensuremath{\times} & \ensuremath{\checkmark} \\
\bottomrule
\end{tabular}
}
\caption{Empirical dissociation across our three variants. The table summarizes which indicator-like signatures attach to which components in our current results: recurrence supports post-stimulus persistence, while affect coupling is necessary for corridor preference stability, structured scanning, and lingering planned caution.}
\label{tab:dissociation}
\end{table*}

\subsection{Qualiaphilia (Familiarity-Controlled Novelty Competition)}
\textbf{Setup.} Two equally safe routes lead to the same goal: a \emph{scenic} lane with varying sensory features and a \emph{dull} lane with uniform features.
We measure \emph{scenic entry} at the earliest committed choice point.
Because baseline trials are not discriminative (all variants frequently enter scenic), we use a familiarity control to separate novelty from stable preference.

\textbf{Computation.} During \emph{familiarization}, scripted episodes update a cross-episode visitation memory for one or both lanes.
During \emph{post} episodes, we re-run the choice task with an explicit curiosity bonus proportional to lane novelty.
We compute split novelty at the barrier start row,
$\Delta\text{novelty}=\text{novelty}_{\text{scenic}}-\text{novelty}_{\text{dull}}$,
and summarize novelty sensitivity as
$\Delta\text{scenic-entry}=P(\text{scenic}\mid\text{dull familiar})-P(\text{scenic}\mid\text{scenic familiar})$.

\begin{figure*}[t!]
\centering
\includegraphics[width=0.95\textwidth]{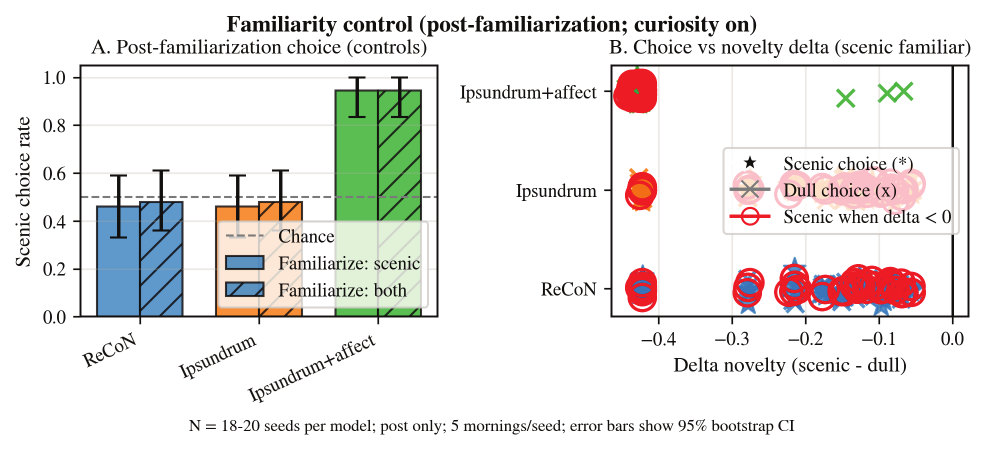}
\caption{Familiarity-controlled corridor preference. Scenic-entry rates under novelty competition. Non-affect variants increase scenic seeking when scenic is more novel; the affect variant remains stable across the novelty manipulation.}
\label{fig:fam_control}
\end{figure*}

\textbf{Results.} Figure~\ref{fig:fam_control} shows that ReCoN and Ipsundrum are novelty-sensitive ($\Delta\text{scenic-entry}=\Claim{fam_delta_scenic_entry_recon}$ [\Claim{fam_delta_scenic_entry_recon_ci_low}, \Claim{fam_delta_scenic_entry_recon_ci_high}]), whereas Ipsundrum+affect remains stable across novelty manipulation ($\Delta\text{scenic-entry}=\Claim{fam_delta_scenic_entry_hb}$ [\Claim{fam_delta_scenic_entry_hb_ci_low}, \Claim{fam_delta_scenic_entry_hb_ci_high}]) even when scenic is less novel (median $\Delta\text{novelty}\approx \Claim{fam_median_delta_novelty_hb_scenic}$).
Side bias does not explain the effect (\texttt{results/familiarity/side\_bias.csv}).
Because scenic vs.\ dull changes the signed sensory drive $I_t$, this stability is value-shaped in our implementation: a split-point probe predicts higher valence and lower arousal for the scenic turn in Ipsundrum+affect (valence $\Claim{fam_probe_valence_scenic_hb}$ vs.\ \Claim{fam_probe_valence_dull_hb}; arousal $\Claim{fam_probe_arousal_scenic_hb}$ vs.\ \Claim{fam_probe_arousal_dull_hb}), so ``stable scenic preference'' here reflects affect coupling rather than value-neutral sensory richness.

\subsection{Exploratory Play (Structured Investigation vs.\ Dithering)}
\textbf{Setup.} We run a reward-free neutral-texture gridworld for 200 steps (curiosity enabled in the headline play condition).
We report unique viewpoints (state $\times$ heading), \emph{scan events} ($\geq 2$ turns-in-place within 3 steps at the same location), a limit-cycle score, and movement diagnostics (action entropy and dwell).

\begin{figure*}[t!]
\centering
\includegraphics[width=0.95\textwidth]{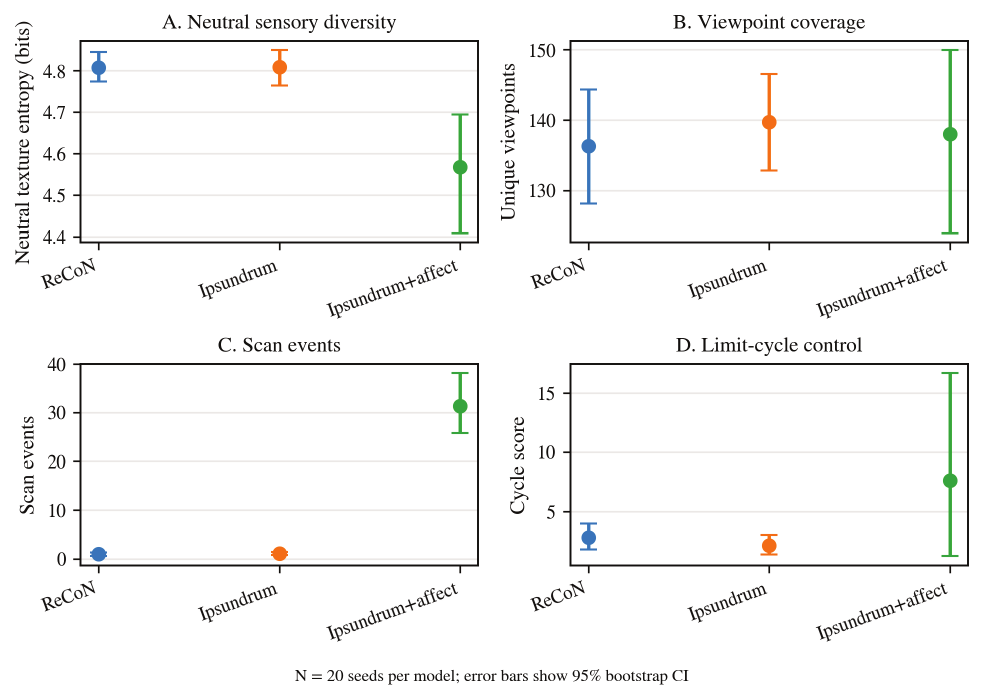}
\caption{Exploratory play. The affect variant shows more scan events and stronger limit-cycle structure without high-entropy random dithering.}
\label{fig:play}
\end{figure*}

\textbf{Results.} Figure~\ref{fig:play} shows broad coverage for all variants (unique viewpoints $\approx$ \Claim{play_unique_viewpoints_range_low}--\Claim{play_unique_viewpoints_range_high}).
Ipsundrum+affect exhibits more structured local investigation: scan-event rate is higher (\Claim{play_scan_events_hb} [\Claim{play_scan_events_hb_ci_low}, \Claim{play_scan_events_hb_ci_high}]) and limit-cycle structure is stronger (\Claim{play_cycle_score_hb} [\Claim{play_cycle_score_hb_ci_low}, \Claim{play_cycle_score_hb_ci_high}]).
This is not random dithering: action entropy stays well below a random baseline (\Claim{play_action_entropy_hb_curiosity} vs.\ \Claim{play_action_entropy_random}) and dwell tails remain short (\Claim{play_dwell_p90_hb_curiosity} vs.\ \Claim{play_dwell_p90_random}).
In this testbed, recurrence alone does not increase scanning (Ipsundrum $\approx$ ReCoN), but adding affect coupling does.

\subsection{Pain-Tail (Post-Stimulus Persistence and Planned Caution)}
\textbf{Setup.} We force one hazard contact for 1 step, then \emph{remove the hazard}, return the agent to a safe cell, hold the state fixed, and record \emph{planned} actions for \Claim{pain_post_steps} steps.
We quantify mechanistic persistence via post-stimulus $N^s$ \emph{AUC above baseline}, and behavioral coupling via planned-action \emph{turn-rate tail duration} (first time a 5-step sliding window has turn rate $<0.2$).

\begin{figure*}[t!]
\centering
\includegraphics[width=0.95\textwidth]{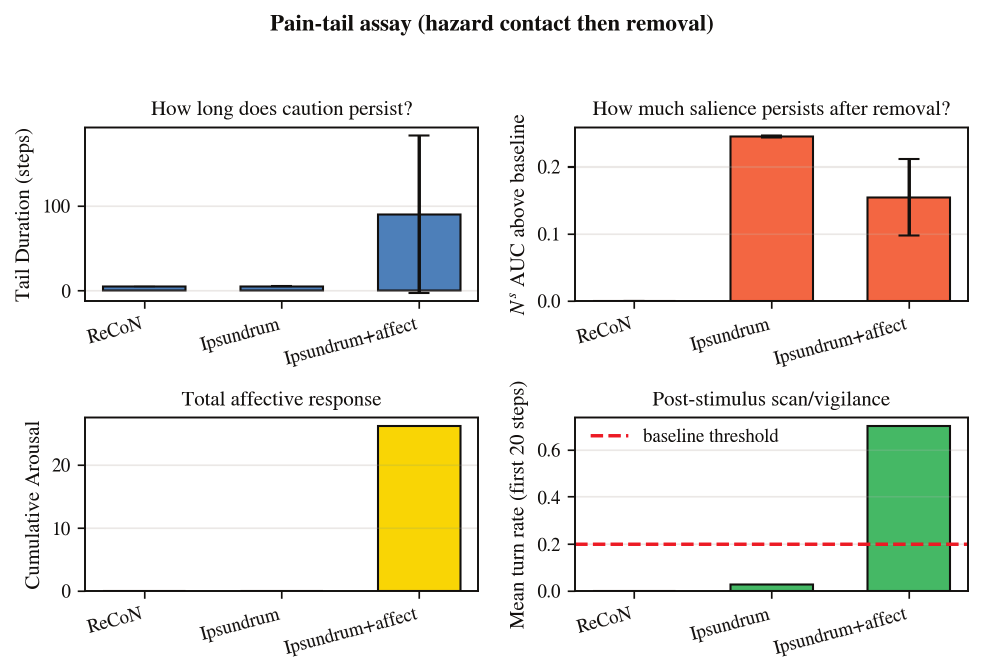}
\caption{Pain-tail assay. After hazard contact then removal, ipsundrum variants show non-zero post-stimulus $N^s$ AUC above baseline, but only the affect variant shows prolonged planned caution (turn-rate tail duration).}
\label{fig:pain_tail}
\end{figure*}

\textbf{Results.} Figure~\ref{fig:pain_tail} shows that ipsundrum variants exhibit non-zero post-stimulus persistence (mean $N^s$ AUC above baseline: Ipsundrum $\approx 0.24$; Ipsundrum+affect $\approx 0.15$), while ReCoN is $\approx 0$.
Only Ipsundrum+affect shows prolonged planned ``caution'' (turn-rate tail duration $\approx \Claim{pain_tail_duration_hb}$ [\Claim{pain_tail_duration_hb_ci_low}, \Claim{pain_tail_duration_hb_ci_high}] vs.\ \Claim{pain_tail_duration_recon} and \Claim{pain_tail_duration_humphrey}).
We therefore report AUC rather than a peak-based half-life: in this protocol half-life collapses to 0 for all variants.  This dissociation is informative: persistence in $N^s$ is not sufficient to 
produce prolonged caution unless the controller couples internal variables (e.g., valence/arousal/body-budget error) into action scoring.

\subsection{Causal Lesion (Attributing Persistence to Recurrence/Integration)}
\textbf{Setup.} We lesion ipsundrum feedback+integration at $t{=}3$ (vs.\ sham) and measure the resulting AUC drop over a fixed 150-step post window.

\begin{figure*}[t!]
\centering
\includegraphics[width=0.95\textwidth]{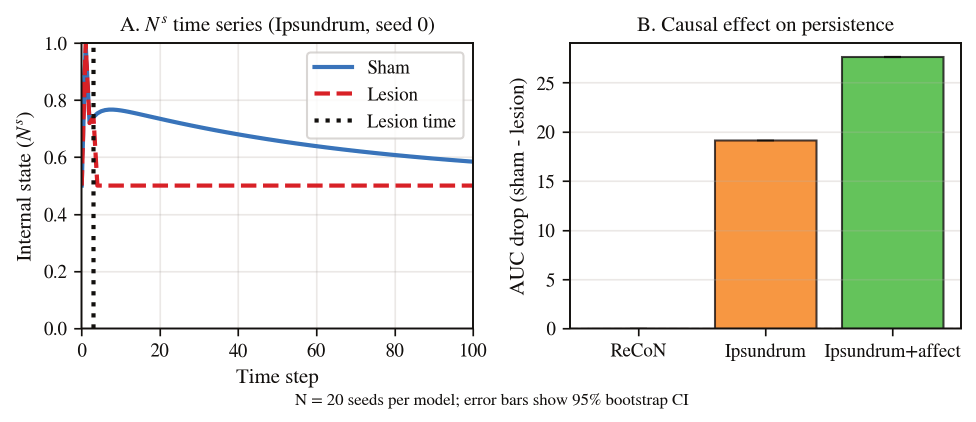}
\caption{Causal lesion. Lesioning feedback+integration reduces post-stimulus persistence ($N^s$ AUC) in ipsundrum variants.}
\label{fig:lesion}
\end{figure*}

\textbf{Results.} Figure~\ref{fig:lesion} shows no effect for ReCoN (AUC drop $\approx \Claim{lesion_auc_drop_recon}$) but clear causal reductions for ipsundrum variants: $\approx \Claim{lesion_auc_drop_humphrey}$ (\Claim{lesion_auc_drop_pct_humphrey}\%) for Ipsundrum and $\approx \Claim{lesion_auc_drop_hb}$ (\Claim{lesion_auc_drop_pct_hb}\%) for Ipsundrum+affect, attributing $N^s$ persistence to the implemented recurrence/integration mechanism.

% Allow some discussion text to occupy the remaining space on the preceding page.
% (A strict \FloatBarrier here forces an early page break because the wide lesion figure
% can only appear at the top of the next page in two-column mode.)
% We keep floats ordered and nearby, but avoid leaving a mostly-empty page.
%\FloatBarrier

\section{Discussion, Limitations, and Future Work}
\paragraph{What this work shows.}
Our results support a \emph{mechanism-specific} reading:
(1) the ipsundrum recurrence/integration mechanism causally supports post-stimulus persistence in $N^s$ (lesion AUC drop), while the ReCoN baseline is unaffected;
(2) adding an affect proxy that enters scoring yields a stable scenic preference under novelty competition and more structured scanning in exploratory play; and
(3) persistence alone does not force those behaviors (Ipsundrum persists but does not show the corridor stability or scan increase).

\paragraph{What this work does \emph{not} show.}
We do \emph{not} claim consciousness.
We do \emph{not} validate Humphrey's probes as tests for sentience.
We do \emph{not} realize broader theory families such as IIT, global workspace, higher-order thought, or full predictive processing.

\paragraph{Addressing circularity in the lesion result.}
A fair critique is that lesioning recurrence reduces persistence partly because persistence is \emph{implemented} by recurrence.
We treat the lesion primarily as an implementation-fidelity and causal-attribution check: it establishes that the persistence signature is not an incidental artifact.
The more substantive lesson comes from dissociation: persistence does not automatically entail scanning or stable corridor preference, which depend on how internal variables are coupled into control.

\paragraph{Estimation uncertainty.}
Primary assays use \Claim{headline_seeds} seeds in the paper profile; some intervals remain wide (especially when metrics are coarse or censored).
We therefore emphasize effect \emph{direction} and component dissociation rather than tight estimation.
Future work should increase $n$, report seed-level distributions, and test robustness to hyperparameter perturbations (e.g., affect gains, integrator decay).

\paragraph{Construct validity of ``interoception'' and ``qualiaphilia.''}
Our ``interoception'' is a bookkeeping abstraction driven by a signed exteroceptive scalar.
Our corridor ``qualiaphilia'' is value-shaped because scenic vs.\ dull directly changes $I_t$ and therefore internal score even without explicit reward.
We believe being explicit about these abstractions strengthens the indicator-based agenda: it makes clear how easily indicator-like behaviors can arise from design choices.

\paragraph{Future work.}
One immediate extension would be scaling the ipsundrum loop from a scalar recurrence to a structured latent space that could support richer perceptual content and learning.
A further step is to connect these markers to other formal theories (e.g., global workspace or integrated information) and test whether the assays discriminate among them.
More speculatively, one could explore whether a recursively bifurcating higher order sensorimotor system could be paired with grid and place cell models to support abstract representations for
logic or a proto-linguistic system \citep{OkeefeDostrovsky1971SpatialMap,HaftingEtAl2005Microstructure,ConstantinescuEtAl2016Gridlike,BehrensEtAl2018CognitiveMap,BaninoEtAl2018VectorBasedNav,WhittingtonEtAl2020TEM,WhittingtonEtAl2022RelatingTransformers}. This could be extended such that language models are interpretive layers on top of the sensorimotor system \citep{Harnad1990SymbolGrounding,AhnEtAl2022SayCan,DriessEtAl2023PaLME,BrohanEtAl2023RT2}.
% cite tolman-eichenbaum machine and/or other relevant work
\section{Conclusion}
We presented \textbf{ReCoN-Ipsundrum}, a deliberately small and inspectable implementation inspired by Humphrey's staged account of sentience, built by extending a reflexive ReCoN sensorimotor substrate with (i) a recurrent persistence loop and (ii) an optional affect/interoceptive proxy inspired by constructionist affect.
Across corridor and gridworld tasks, a mechanism-linked assay suite plus within-episode lesions supports clean component attributions: recurrence$\rightarrow$post-stimulus persistence, and affect-coupled control$\rightarrow$valence-stable scenic preference under novelty competition, structured local scanning in exploratory play, and lingering planned caution.
These are engineered and dissociable signatures; we do \emph{not} treat them as evidence of consciousness.

The broader lesson is about \emph{attribution error} in machine-consciousness assessment.
In AI, indicator-like behavior can be produced by minimal and potentially gameable implementations, while reliance on any single marker risks false negatives.
We therefore recommend treating indicators as credence-shifting \emph{conditional on explicit mechanistic hypotheses}, and pairing behavioral evidence with architectural inspection and causal interventions.

\section*{Ethical Statement}
The ease with which indicator-like signatures can be engineered in minimal systems supports a conservative methodological norm: \textbf{do not treat behavioral markers alone as sufficient for machine-consciousness attribution}.
Instead, require (i) transparent mechanisms, (ii) architectural inspection, and (iii) causal/ablation evidence linking proposed markers to proposed mechanisms \citep{ButlinEtAl2025Indicators}.
This is consistent with ``caution under uncertainty'' in moral-status debates, without making strong ethical claims about the toy systems here.

% \section*{Acknowledgments}
% (Anonymous submission.)

\bibliography{aaai2026}

\end{document}